# Anomaly Triplet-Net: Progress Recognition Model Using Deep Metric Learning Considering Occlusion for Manual Assembly Work

Takumi Kitsukawa[a], Kazuma Miura[a] Shigeki Yumoto[a] Sarthak Pathak[a] Alessandro Moro[b] and Kazunori Umeda[a]

[a]Chuo University, 1-13-27 Kasuga, Bunkyo-ku, Tokyo, Japan ; [b]RITECS Inc., 3-5-11 Shibasaki, Tachikawa-shi, Tokyo, Japan

**ABSTRACT**
In this paper, a progress recognition method consider occlusion using deep metric learning is proposed to visualize the product assembly process in a factory. First, the target assembly product is detected from images acquired from a fixed-point camera installed in the factory using a deep learning-based object detection method. Next, the detection area is cropped from the image. Finally, by using a classification method based on deep metric learning on the cropped image, the progress of the product assembly work is estimated as a rough progress step. As a specific progress estimation model, we propose an Anomaly Triplet-Net that adds anomaly samples to Triplet Loss for progress estimation considering occlusion. In experiments, an 82.9 [%] success rate is achieved for the progress estimation method using Anomaly Triplet-Net. We also experimented with the practicality of the sequence of detection, cropping, and progression estimation, and confirmed the effectiveness of the overall system.

**KEYWORDS**
deep metric learning; triplet loss; occlusion; progress recognition; assembly;

## 1. Introduction

In recent years, shortage of labor in the manufacturing industry has become increasingly serious. Therefore, there is a need to improve productivity by increasing work efficiency on the production line. As a solution, automation of production lines using IoT and robots is being promoted. In particular, research on smart manufacturing is being promoted actively [1,2]. However, automation has not progressed in high-mix, low-volume factories that produce products according to customer needs, and the majority of work is carried out manually. Such factories have not yet been converted to smart factories due to the lack of automated data capture. In particular, product assembly work requires changes to the production line each time product specifications change if automation is used. Therefore, the work is often performed manually, making automated measurement difficult. In addition, it is not desirable to install new measurement sensors at the site due to cost and labor considerations.

CONTACT Takumi Kitsukawa. Email: kitsukawa@sensor.mech.chuo-u.ac.jp
CONTACT Kazuma Miura. Email: miura@sensor.mech.chuo-u.ac.jp

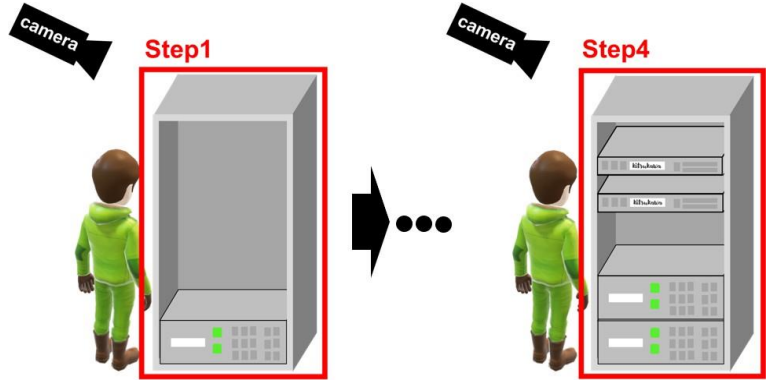

**Figure 1.** Proposed method expected result

In response to this, it is considered effective to introduce a system that automatically manages work by acquiring images from fixed-point cameras installed in factories and estimating the progress of the work. There are research on motion recognition using skeletal point information of a person in a time series [3,4], but the assembly of large equipment often requires several days of work and the assembly sequence is not defined in detail, so progress estimation using worker movements and time series data is not suitable. In addition, because of the movements might change from person-to-person, using skeletal point information is not suitable.

Although assembly progress estimation methods focused on the object have been proposed [5], they have not been able to deal with occlusion by the worker. In fact, in experiments, misjudgment occurred when the worker was overlapped with the object, resulting in reduced accuracy.

The aim of this research is to construct a system that estimates the progress of assembly work by focusing on the objects to be assembled consider occlusion. An overview of the proposed system is shown in Figure 1. Progress estimation uses a method that judges which step has been reached in relation to the assembly progress step set earlier.

## 2. Use of Deep Metric Learning

To construct a system that estimates which step currently belongs to a given step, the problem can be set up as a class classification problem. Deep learning methods using CNN (Convolutional Neural Networks) are highly accurate for classifying classes in images. However, the class classification in this assembly progress estimation is considered to have small differences in appearance between classes, and it is required to extract detailed feature differences and to have as much separation as possible between steps that are far apart. Therefore, the use of metric learning [6], one of the machine learning methods, is considered.

Metric learning is a method for learning patterns that transforms input data into a feature space so that the samples are separated into classes based on the distance or similarity between them. The aim of metric learning is to increase the distance between samples of different classes while decreasing the distance between samples of the same class. Figure 2 shows motion in feature space for metric learning. In particular, deep metric learning [7], which uses a multi-layer neural network to extract features when learning patterns, has also been proposed.

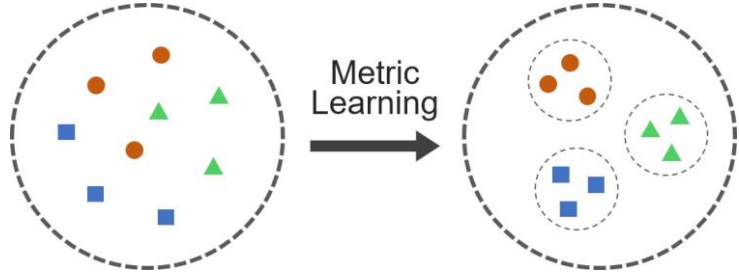

**Figure 2.** Motion in feature space for metric learning

In general class classification, the features extracted by the CNN are passed through Fully Connected layer and converted into class affiliation probabilities using the SoftMax function. However, this means that the feature extraction network is trained without considering the distance between samples of the same class and other classes. Deep metric learning, on the other hand, can obtain discriminative features by deliberately increasing the distance between samples of other classes and decreasing the distance between samples of the same class. Therefore, it performs well when the samples of each class are small or when there are unknown classes.

Considering the practical application in an actual factory, the assembly work progress estimation using our proposed method will require only a small amount of data to learn from. Therefore, we propose the use of deep metric learning, which can learn feature differences even with small amount of data.

## 3. Proposed Method

### 3.1. Overview of the Proposed System

An overview of the proposed system is shown in Figure 3. In the initial phase, an object detection model is trained on a custom dataset to be able to detect the position of the object to be assembled in the image. In addition, a step estimation model is trained to estimate the progress of the cropped image. The flow of the system is to detect objects in the image acquired from a fixed-point camera using an object detection method, crop their positions and estimate their progress using a step estimation model based on deep metric learning.

### 3.2. Step Estimation Method

The structure of the proposed step estimation method is shown in Figure 4. First, the steps to be judged in the assembly progress are set and training data are prepared. The method is to cut out the part of the object from the assembly video and save it separately for each step. One image is randomly selected from the training data and an anchor sample is set. The positive sample is the image from the same step as the anchor sample, and the negative sample is the image from a different step. In the example in Figure 4, the image from step 5 is set as the anchor sample. Another image from step 5 is selected as a positive sample, and an image from a different class than the anchor sample, step 0, is selected as a negative sample. Next, these three images are input to a four-layer CNN model and then to a one-layer all-unions layer to obtain a 128-dimensional feature vector. The weights of the CNN models are shared. The parameters are updated by defining a loss function to increase the distance between the anchor sample and the positive sample and to decrease the distance between the anchor sample and the negative sample among the three feature vectors obtained. Stochastic Gradient Descent is used in the optimization algorithm to learn by mini batch.

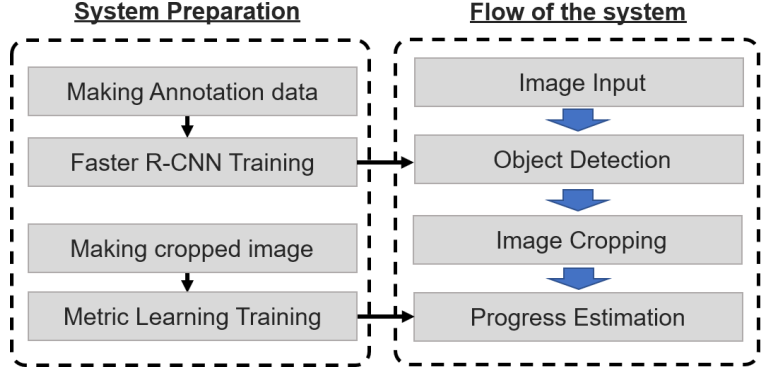

**Figure 3.** Flow of the proposed system

The loss function uses Triplet Loss [10], which calculates the relative distance between the anchor, positive and negative samples as a set of data. The equation is shown in (1).

$$L_{Triplet} = max(d_p - d_n + m, 0) \qquad (1)$$

where $d_p$ is the distance between the anchor sample and the positive sample in the feature space and $d_n$ is the distance between the anchor sample and the negative sample. $m$ is an arbitrary constant that represents the degree of movement of the distance away/near work, called the margin. The Euclidean distance in the feature space is used to calculate the distance. In other words, by reducing this loss function, features of images at the same step can be brought closer together in feature space and features of images at different steps can be moved away from each other in feature space.

Next, the learned CNN model is used to embed each image to the metric learning feature space. First, the images used for each step in the training are input again one by one into the trained model and embedded in the feature space. Once all the images have been converted into feature vectors, the input images to be estimated are also input into the learned model and embedded in the feature space. For these embedded data, the k-Nearest Neighbor algorithm [8][9] is used to estimate which step the unknown test data belongs to. Furthermore, by judging errors as those whose distance from all the clusters in the feature space in the feature space is farther than a threshold value, it is possible to detect assembly errors in assembly work and to reduce misjudgments that occur in occlusion.

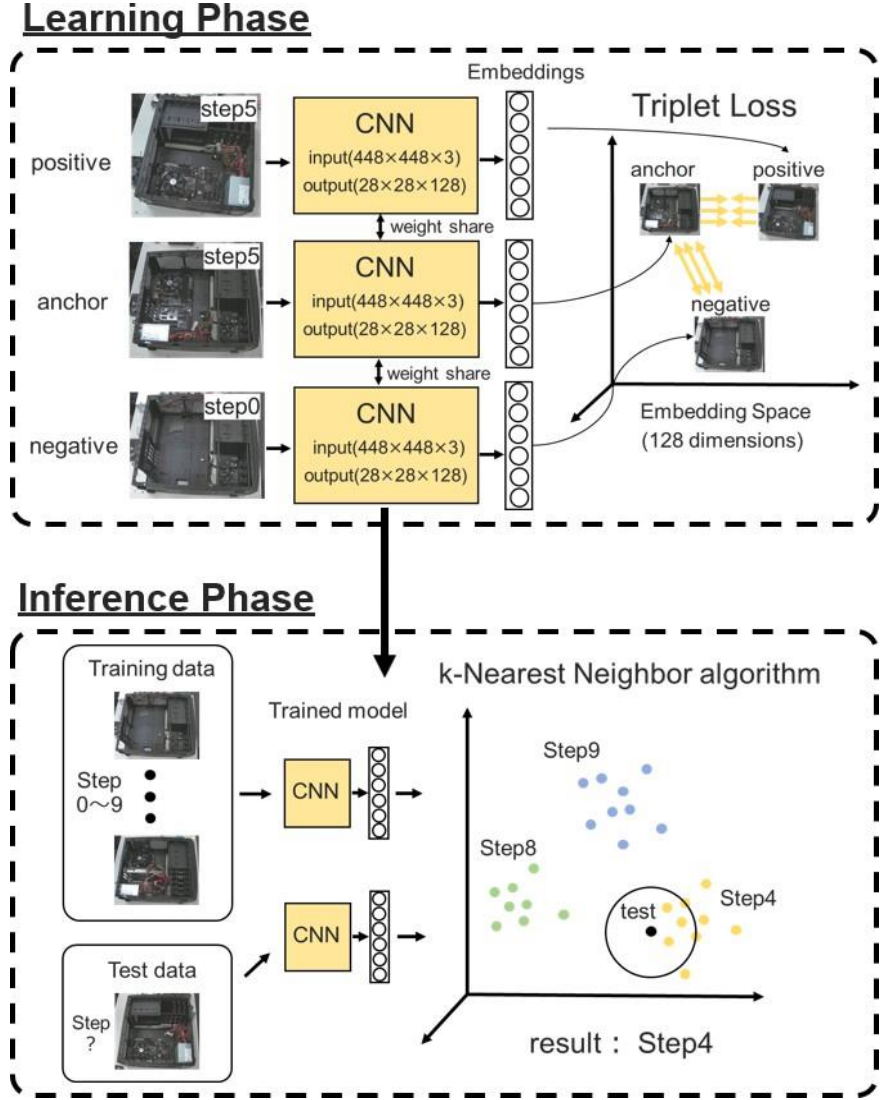

**Figure 4.** Network structure of the proposed step estimation method

### 3.3. *Adaptive Triplet Loss: Reduces mistakes with neighboring steps*

The estimation of progress steps in assembly operations is considered to be prone to misjudgment of close steps due to the similar features of neighboring steps. Therefore, we propose a method to adaptively change the value of the margin of Triplet Loss to be used so that it becomes larger when the steps are close. The formula uses a normal distribution so that the margin falls smoothly from nearby to faraway steps. The proposed adaptive margin formula is shown in (2).

$$m = \sqrt{\frac{1}{2\pi\sigma}} \exp\left(-\frac{(n_n - n_a)^2}{2\sigma^2}\right) \cdot a \qquad (2)$$

where $\sigma^2$ is the variance of the number of steps, $a$ is an arbitrary constant repre-senting the size of the margin, $n_a$ is the number of steps in the anchor sample and $n_n$ is the number of steps in the negative sample. The examples of adaptive margins when learning steps 1 and 5 as anchors are shown in Fig. 5.

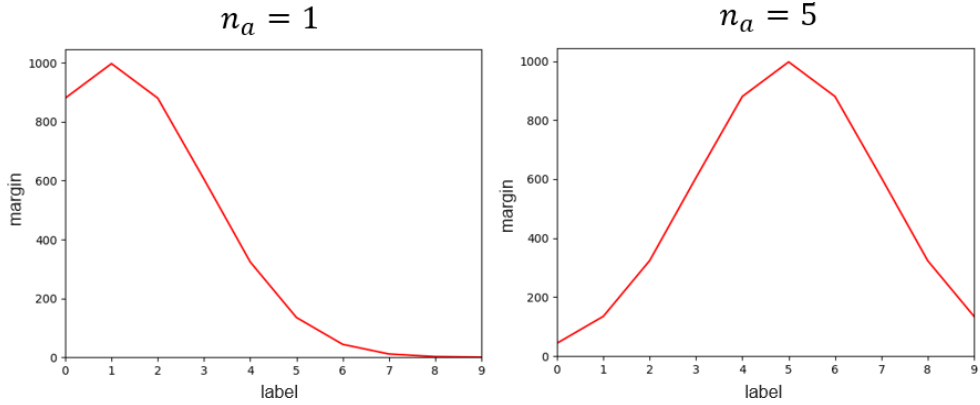

**Figure 5.** Examples of adaptive margins in learning step 1 and 5

### *3.4. Anomaly Triplet-Net: Models that consider occlusion*

In actual assembly operations, occlusion caused by various obstacles such as workers is likely to occur. Therefore, we propose Anomaly Triplet-Net, which can recognize images that cannot be estimated due to occlusion as abnormal images without including abnormal data in the training data.

*3.4.1. Model structure*

Figure 6 shows the model structure of the Anomaly Triplet-Net learning model that takes into account anomalies. The images are then input to the CNN layer in the same manner as the three samples and converted into 128-dimensional feature data. By learning the positional relationship of the four obtained data in the feature space, estimation can be performed taking into account abnormal images.

Random Erasing is used to mask some short penalty areas in the image with random pixel values. Figure 7 shows an image actually noised by Random Erasing. Noise is applied to some or most of the objects in the image, making it impossible to correctly determine the step.

The parameters of Random Erasing are set to randomly select a range of 0.3 to 0.5 for the ratio of the rectangle area to the input image area, and a range of 0.3 to 3.3 for the aspect ratio of the rectangle drawn on the image. Random Erasing is repeated twice for each image to accommodate different occlusions.

*3.4.2. Proposed loss function*

This section describes the loss function of the proposed Anomaly Triplet-Net method considering occlusion.

Figure 8 shows the positional relationship in the feature space. In addition to the positional relationship of each sample in the conventional method, Triplet Loss, an additional "anonymous" sample is added. The proposed loss function is shown in (3).

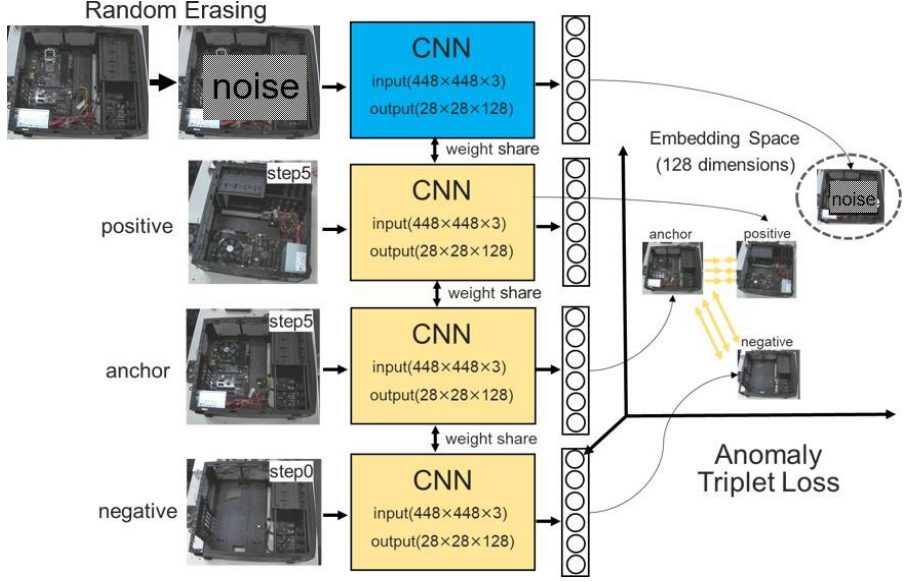

**Figure 6.** Network structure of the Anomaly Triplet-Net

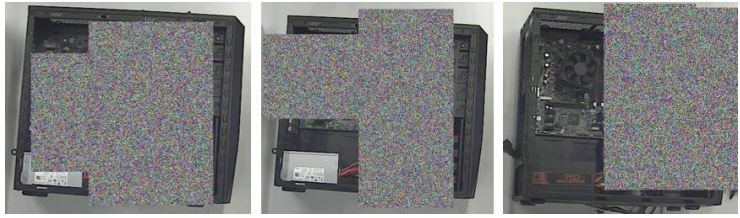

**Figure 7.** Anomaly sample generated with noise by Random Erasing

$$L_{AnomalyTriplet} = max(d_p - d_n + m_\alpha, 0) + \lambda \cdot max(d_n - d_{ano} + m_\beta, 0) \quad (3)$$

$m_\alpha$ and $m_\beta$ are the margins of each term, and $d_{ano}$ is the distance between the center of gravity of the anchor, positive, and negative samples, respectively, and the distance between the center of gravity of the anomaly sample. $\lambda$ is a coefficient that adjusts the weights of the $d_p$ and $d_n$ terms and the $d_n$ and $d_{ano}$ terms.

It is desirable to learn only the positions of the anchor, positive, and negative samples at the beginning of the learning process, and to learn the distance from the anomaly sample as the learning process progresses. Therefore, the coefficient $\lambda$, which adjusts the weight of each term, is not constant, but is a variable corresponding to the number of training sessions.

### *3.5. Object Detection*

For object detection and training data preparation, we will use Faster-R CNN [11] and Siam Mask [12], which have been validated in our research [13].

The Faster-R CNN used in object detection has a model structure that discrimi-nates whether the contents of a rectangle are objects or background and classifies the detected regions. Unlike conventional object detection methods such as R-CNN [14] and Fast R-CNN [15], it extracts candidate object regions using a CNN (Convolutional Neural Network) structure called RPN (Region Proposal Network) [16], which can significantly reduce processing time. Object detection is enabled by fine-tuning this pre-trained model of Faster R-CNN with a custom dataset created for assembly products.

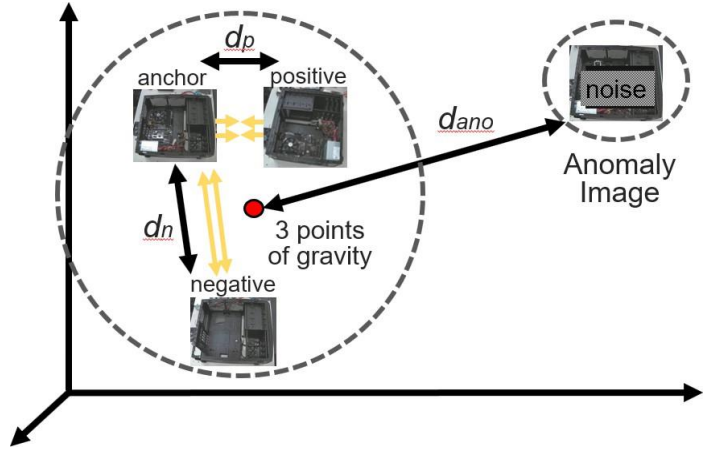

**Figure 8.** Location of each sample including anomaly sample in the feature space

Siam Mask, used to create training data, is a mask-based object tracking method. This tracking method allows for continuous labeling of the video for training. Specifically, given the position of an object to be tracked in the first frame of the video, it will estimate the object's position in all subsequent frames. This allows for efficient creation of training datasets.

## 4. Experiments

Experiments were conducted to evaluate the proposed step estimation model and the system as a whole. A desktop PC was used as the object of the assembly product. The camera was an IO DATA network camera " Qwatch TSWR-LP "[17] and the pixel value was set to 1980 × 1080. Two cameras were set up in order to get more training data in a one-time assembly operation. The positions of the two set up cameras and the actual images captured are shown in Figure 9. The camera position is about 1.8 m away from the object and the angle is about 90 ° down from the horizontal. Our proposed method requires that the steps of the progression be set up ahead of time by yourself. In the desktop PC assembly experiment, the steps to be judged were the 8 steps from step 1 to step 8. The steps are shown in Figure 10.

We used a total of 320 images for training, 40 at each step. To prepare the dataset, the location of the target desktop PC was cut out from each image and stored in separate folders for each step. The image size and number of convolution layers of the deep metric learning model to be implemented is four layers of the proposed model. The constant k of the k-nearest neighbour method used to judge step was set to k = 10. We determined k = 10 because the number of training data to be input into the trained model in the progress estimation phase is 40 per step.

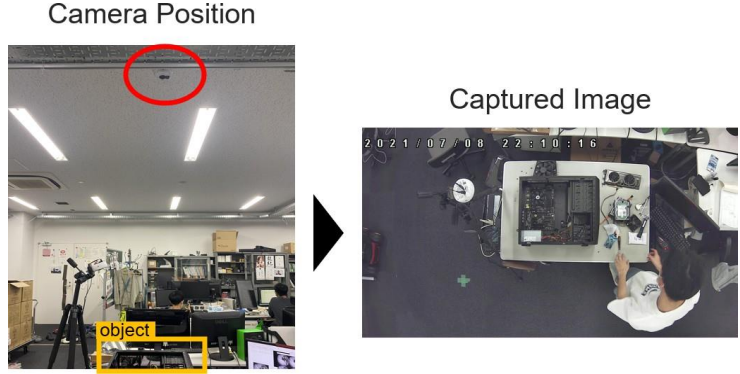

**Figure 9.** Camera position and captured images

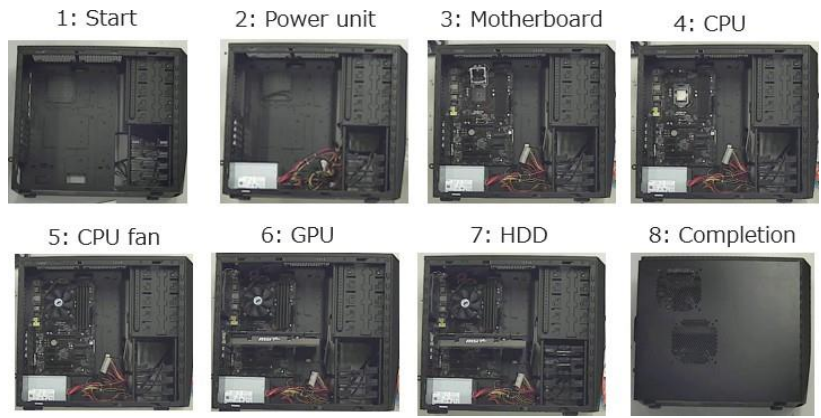

**Figure 10.** The point of dividing steps

In addition, as this study assumes a small amount of training data, the data was expanded by randomly adding rotation, projective transformation, colour change and partial truncation to the input images in order to avoid data bias. The actual images processed for data expansion are shown in Figure 11.

### 4.1. Implementation of the proposed method

The model was implemented based on the experimental conditions.

For the reasons explained in 3.4.2, the coefficient $\lambda$, which adjusts the weight of each term, is set to 0 until 50 epochs, and $\lambda$ is set to gradually increase after 50 epochs.

Figure 12 shows the distance between the anchor sample and the positive sample, the distance between the anchor sample and the negative sample, and the distance between the center of gravity of the three points (anchor, positive, and negative) and the anonymous sample, respectively, in the feature space.

Figure 12 shows that the distance between $d_n$ and $d_p$ increases as the number of learning cycles increases. In addition, the distance between $d_{ano}$ and both $d_p$ and $d_n$ increases after 50 epochs, when $d_{ano}$ begins to have the weight of the loss function term. From these results, we can say that the proposed model and the loss function

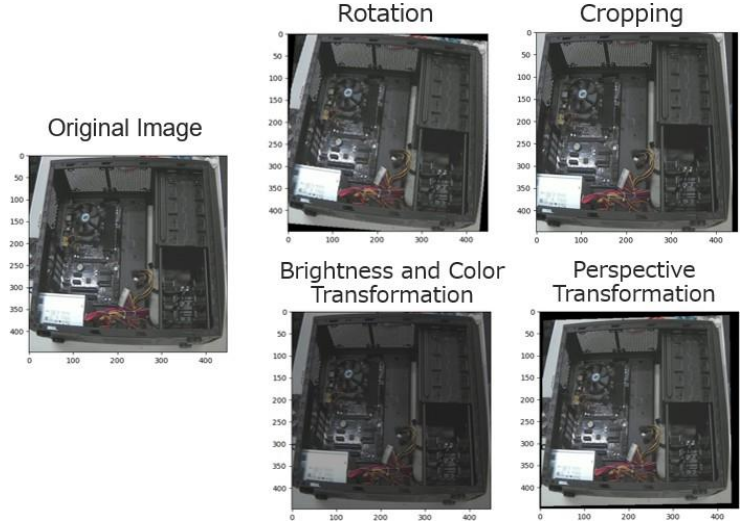

**Figure 11.** Image transformation for data augmentation

correctly learn anomaly sample in the feature space.

## 4.2. Experiments on Step Estimation

Figure 13 shows the confusion matrix of the proposed method with 100 epoch training, and Figure 14 shows the results of 2D visualization of the feature space using t-SNE [18].

We also compared the Adaptive Margin method, which is a margin suitable for progress estimation, with Anomaly Triplet-Net, a learning model that considers occlusion. Table 4.2 shows the results of the percentage of correct responses for the general Triplet Loss method using fixed margin and Adaptive Margin, and for the proposed Anomaly Triplet-Net method using fixed margin and Adaptive Margin.

The method using Adaptive Margin for the proposed Anomaly Triplet-Net achieved the highest percentage of correct answers (82.9%). The confusion matrix confirms that the proposed method succeeds in estimating with high accuracy.

Furthermore, the results of two-dimensional visualization by t-SNE showed that the samples of the same step were coherent, and that the distance between the samples with anomalies was far from the anomaly sample. These results verify the effectiveness of the proposed method.

## 4.3. Experiments on the Whole System

### 4.3.1. Experimental conditions

Experiments were conducted on a sequence of object detection, cropping, and progress estimation.

For object detection, we used Siam Mask, an object tracking method, to label approximately 1000 images, and fine-tuned a Faster R-CNN. The Faster R-CNN was implemented in Detectron2 [19], a library for deep learning, and fine-tuned on a model that had been previously trained on the MS COCO dataset [20]. To avoid misjudg-ments caused by worker occlusion, we used a method that determines the definitive step when the same decision is made in five consecutive frames.

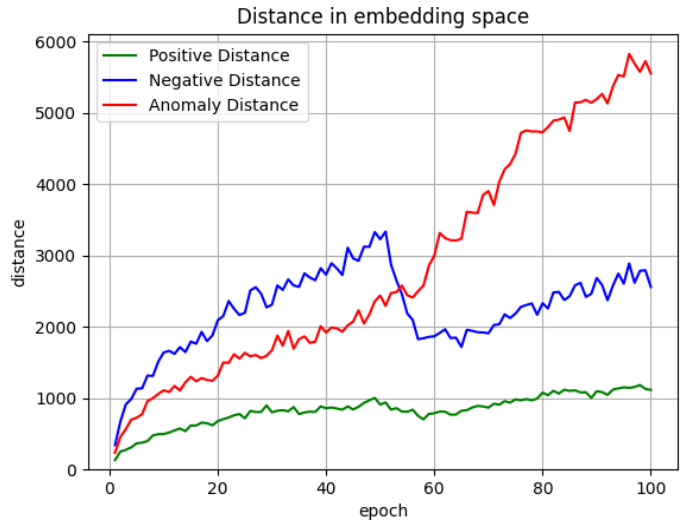

**Figure 12.** Distance of each sample in the feature space in number of training

**Table 1.** Comparison of accuracy with each method

| Loss Function Name | Accuracy |
|---|---|
| Triplet (Fixed) | 0.793 |
| Triplet (Adaptive) | 0.796 |
| AnomalyTriplet-Net (Fixed) | 0.826 |
| AnomalyTriplet-Net (Adaptive) | 0.829 |

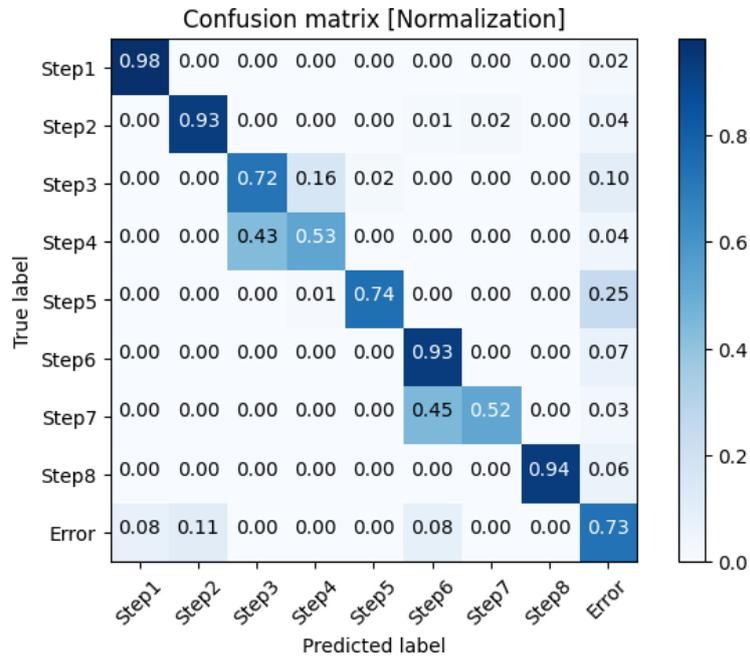

**Figure 13.** Confusion matrices at the predicted and true steps

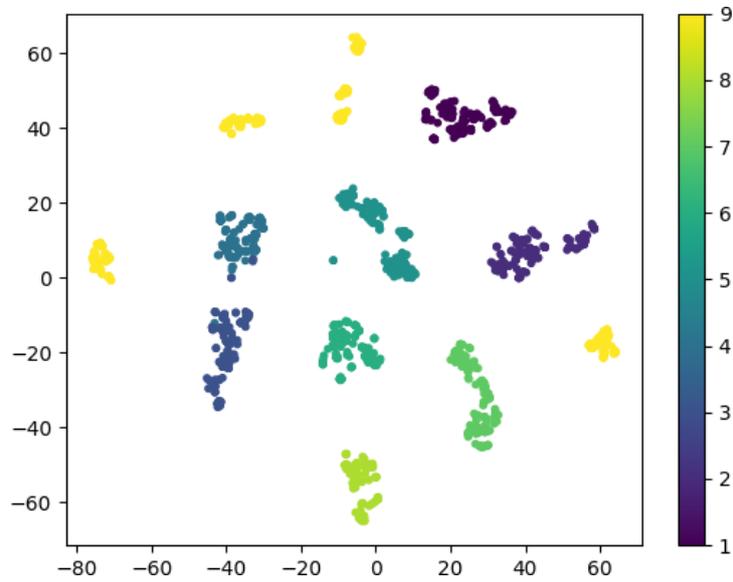

**Figure 14.** Visualization of feature space using t-SNE

*4.3.2. Experimental Results and Discussion*

As a result of the experiment, correct decisions were logged at all steps. Figure 15 shows the actual system in action. The yellow bounding box indicates the detection of an object, and the red numbers on the right side indicate the steps that were judged.
As a quantitative evaluation, Figure 16 shows a graph with the time of the assembly work video on the horizontal axis and the estimated steps on the vertical axis. The black dots indicate the estimated steps, and the red dots indicate the estimated steps that resulted in errors. The effectiveness of the proposed Anomaly Triplet-Net was confirmed.

The method described in the previous section, in which a step is determined when the same judgment is made for five consecutive frames, was used to definite the step. Figure 17 shows the results of the definitive steps obtained by the method described in the previous section. This graph shows that the number of estimated steps increased by one. However, step4 was not determined correctly and remained at step3. This is because the characteristics of step4 and step3 are very similar. However, in actual use, a rough judgment of steps is required, and if there are only small differences between steps and the misjudgments continue, the problem can be solved by merging them together. Therefore, it can be said that the system is valuable for estimating the assembly progress of large products, which is the subject of this research, even if some misjudgments are made with neighboring steps.

These results also validate the effectiveness of the system as a whole, including object detection.

### *4.4. Evaluation of Processing Speed*

*4.4.1. Experimental conditions*

Considering the practical application of this research, the processing speed of the proposed method was measured.

The PC used for measuring had an Intel Core i7-8700 3.7GHz CPU and NVIDIA GeForce RTX 2080 GPU. The time taken for each processing when the resolution of the input image is 1920 × 1080 and 1280 × 720 is shown in Table 2.

*4.4.2. Experimental Results and Discussion*

Note that this time the processing speed is based on the case where there is only one object. Comparing detection and step estimation, object detection takes about 16 times longer than step estimation. This is thought to be related to the fact that step estimation uses a cropped image as input and the image size is small. Although the processing time is shorter with 1280 × 720, there was no significant change in processing speed due to the difference in input image size. The overall system processing speed was about 5 [fps], which is considered sufficient for real-time measurements.

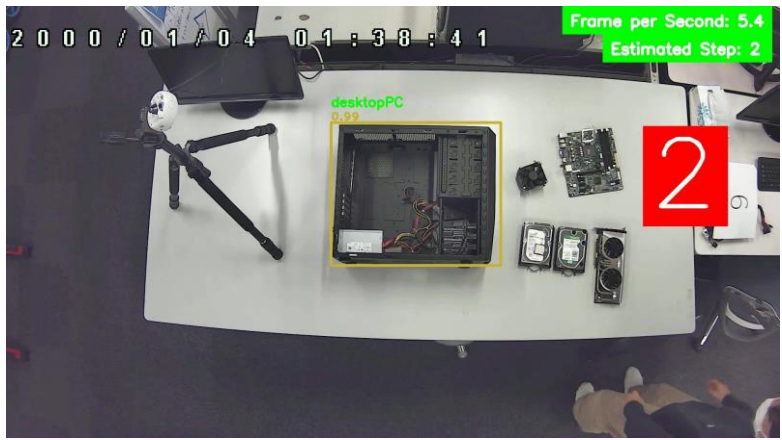

**Figure 15.** Successful object detection and progress estimation

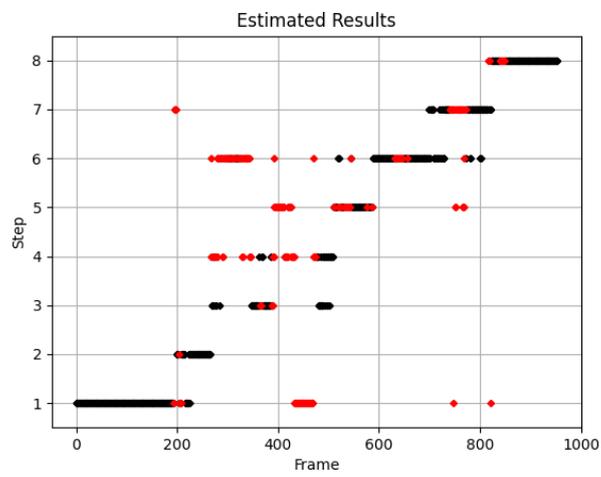

**Figure 16.** Relationship between assembly operation video time and estimated results

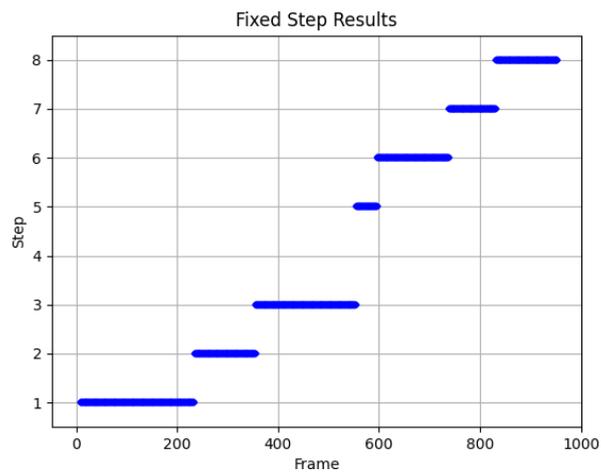

**Figure 17.** Results of the definitive step obtained from the estimation results

Table 2. Processing speed of Proposed system

|       | Detection [ms] | Step Estimation [ms] | Whole System [ms] | Whole System (frame rate)[fps] |
|-------|----------------|----------------------|-------------------|--------------------------------|
| 1080p | 188.4          | 11.6                 | 202.3             | 4.9                            |
| 720p  | 181.4          | 10.8                 | 192.5             | 5.2                            |

## 5. Conclusion

We proposed a progress estimation system consider occlusion for assembly work focusing on objects. Specifically, we proposed a progress estimation method that detects and crops the object and uses deep metric learning to judge the steps. We also proposed Anomaly Triplet-Net based on Triplet Loss to reduce misjudgments when occlusion occurs because of worker or anything. Experiments showed the effectiveness of the method by evaluating the step estimation part and the whole system including detection.

In the experiments on the step estimation part, an 82.9 [%] success rate is achieved for the progress estimation method using Anomaly Triplet-Net. Experiments on the whole system successfully achieved progress estimation in a series of detection, cropping and step estimation. From the quantitative evaluation, it was possible to correctly determine errors when occlusion was occurring, thus reducing the number of false determinations. The processing speed was about 5 [fps], which was sufficient for real-time measurements.

As future work, the system is required recognition between steps. The progress estimation in the form of step judgments used in this study has the problem of not being able to recognize the work in between steps. We believe that the system would be more versatile if it could indicate which step is closer to the neighboring step when recognizing the work in progress between steps. Therefore, we aim to implement a method that can estimate the progress between steps by calculating the distance relationship in the feature space even when the work does not belong to any step.